\documentclass[runningheads]{llncs}

\usepackage{footmisc}

\usepackage{geometry}

\usepackage[
    backend=biber,
    style=numeric-comp, 
    sorting=none,
    maxcitenames=2,
    mincitenames=1,
    autocite=superscript
]{biblatex}
\newcommand{\authorcite}[1]{\citeauthor{#1}\,\supercite{#1}}

\DefineBibliographyStrings{english}{%
    andothers = {et al\adddot}
}

\usepackage{bm}
\usepackage{verbatim} 
\usepackage{hyperref} 

\usepackage[T1]{fontenc}
\usepackage{graphicx}
\usepackage{color}

\begin{document}

\title{ECG-FM: An Open Electrocardiogram Foundation Model}
\titlerunning{ECG-FM: An Open Electrocardiogram Foundation Model}
\author{
    Kaden McKeen\inst{1,2,3,4,5} \and
    Sameer Masood\inst{1,8} \and
    Augustin Toma\inst{4,6} \and
    Barry Rubin\inst{1,2,3} \and
    Bo Wang\inst{1,2,3,4,5,6,7}
}
\authorrunning{McKeen et al.}


\institute{
Toronto General Hospital Research Institute, University Health Network, Toronto, Canada \and
Peter Munk Cardiac Centre, University Health Network, Toronto, Canada \and
UHN AI Hub, University Health Network, Toronto, Canada \and
Vector Institute for Artificial Intelligence, Toronto, Canada \and
Department of Laboratory Medicine and Pathobiology, University of Toronto, Toronto, Canada \and
Department of Medical Biophysics, University of Toronto, Toronto, Canada \and
Department of Computer Science, University of Toronto, Toronto, Canada \and
Department of Medicine, University of Toronto, Toronto, Canada
}

\maketitle

\begin{abstract}
Conventional task-specific electrocardiogram (ECG) analysis models require large annotated datasets to train. Foundation models mitigate this burden by leveraging self-supervised pretraining; however, the scarcity of open-weight ECG foundation models hinders adoption and cross-study comparability. We present ECG-FM, an open foundation model for ECG analysis, and conduct a study using a dataset of 1.5 million ECGs. ECG-FM is a transformer-based model pretrained using a hybrid contrastive and generative self-supervised learning approach. Our downstream tasks include predicting reduced left ventricular ejection fraction (LVEF) and ECG interpretation labels, where we release a benchmark task on the MIMIC-IV-ECG dataset. We affirm that ECG-FM is robust, label-efficient, and functionally discriminative by showcasing data scaling experiments, performing a latent space analysis, and generating saliency maps. ECG-FM markedly outperforms task-specific models in the small-to-medium-scale data regime and demonstrates cross-dataset generalizability, achieving high AUROC on many clinically salient labels such as atrial fibrillation (0.996) and LVEF $\leq 40\%$ (0.929). We release our code, model weights, and benchmark task at \url{https://github.com/bowang-lab/ECG-FM/}.
\end{abstract}

\section{Introduction}

The electrocardiogram (ECG) is a standard, low-cost, non-invasive diagnostic test that has been ubiquitous in the assessment and management of cardiovascular disease for decades. Since human ECG interpretation quality varies greatly according to the level of expertise \supercite{RN938}, computerized interpretations have been used to improve interpretation accuracy and mitigate interobserver variability. However, traditional rule-based methods of ECG interpretation tend to fall short of providing clinically accurate interpretations, warranting caution when used for clinical decision-making \supercite{cielowlyregarded}. \\

Artificial intelligence (AI)-based ECG analysis methods have been shown to outperform traditional computerized interpretation \supercite{RN930, RN910} and can detect patterns largely unrecognizable to humans, thereby transforming the ECG into a precise, non-invasive biomarker. Recent technological advancements and the growing number of large, publicly available datasets have sparked an increase in AI-based methods for the analysis of clinical cardiac data \supercite{RN930, RN911, RN910, RN912, RN933}. Certain models even exhibit performance that matches or exceeds that of humans \supercite{RN943, RN911}, such as \authorcite{ecg-smart}, which outperformed practicing clinicians when identifying myocardial infarction on ECGs. \\

The field of medical AI is experiencing a paradigm shift, wherein conventional task-specific models are being replaced by foundation models. A foundation model is characterized by its rapid adaptability to new tasks, which is generally achieved through transfer learning techniques; namely, being trained on vast amounts of unlabeled data in a process known as self-supervised learning (SSL), or pretraining, such that they can then be finetuned on any number of downstream tasks requiring labeled, task-specific datasets. This leads to improved data efficiency and reduced reliance on costly labeled data. In light of this, foundation models are quickly becoming the standard for generating state-of-the-art models for applications as diverse as medical imaging segmentation \supercite{Ma2024}, diagnostics and disease prediction \supercite{RN911, RN913}, and medical text interpretation \supercite{RN841}. \\

ECG foundation models commonly rely on transfer learning to improve label efficiency, learning to encode ECG structure and semantics prior to encountering annotated data. Their pretraining is typically performed using SSL techniques that fall into either generative or contrastive categories, each possessing characteristic strengths and drawbacks. \\

Generative SSL methods learn to reconstruct signals from partially masked counterparts \supercite{ecgbert, heartbeit, st-mem}, a process that emphasizes encoding local, low-level structural patterns. However, they do not strongly prioritize high-level semantic feature learning \supercite{liu2023improving}, which may cause the model to underrepresent broader patterns indicative of underlying cardiac function. \\

Conversely, contrastive SSL approaches tend to encode semantic information well, generally by learning to discriminate between global representations of augmented ECG samples \supercite{clocs, eldele2021time, wang2023adversarial}. Nevertheless, these methods remain susceptible to faulty alignment—a phenomenon in which augmentations degrade physiologically significant patterns, misalign their semantic meaning, and thus train the model to encode non-representative features \supercite{faultyalignment, st-mem}. \\

Hybrid SSL techniques seek to leverage the benefits of both generative and contrastive learning \supercite{song2024hybrid, wcr}, aiming to capture low-level patterns while preserving high-level semantic information. \authorcite{song2024hybrid} proposed a hybrid SSL approach that combines masking and contrastive objectives, yet its augmentation-driven contrastive strategy remains vulnerable to faulty alignment. \\

\authorcite{wcr} proposed a hybrid approach that circumvents faulty alignment and is based on well-established SSL objectives. They adopted the transformer-based model and generative masking objective from wav2vec 2.0 \supercite{wav2vec2}, as well as the Contrastive Multi-Segment Coding (CMSC) objective, which was originally introduced in a family of ECG-specific contrastive techniques called Contrastive Learning of Cardiac Signals (CLOCS) \supercite{clocs}. CMSC treats temporally adjacent ECG segments as positive pairs, exploiting the relative stability of cardiac function over short intervals and eliminating the need for augmentation altogether. \authorcite{wcr} also introduced Random Lead Masking (RLM), an ECG-specific augmentation wherein leads are stochastically masked, and demonstrated that by exposing the model to diverse lead combinations during pretraining, their model can be finetuned using arbitrary reduced lead sets of the standard 12-lead ECG \supercite{wcr}. This leads to a flexible model which prioritizes both low-level patterns and high-level semantic information, while also addressing common drawbacks associated with SSL approaches. \\

Progress in the ECG analysis field is hindered by unshared codebases and weights, leading to irreproducible findings, diminished methodological comparability, siloed research, and redundant training efforts. This is particularly problematic in the case of SSL methods, which are often prohibitively expensive to train even when code is released. Although data privacy concerns sometimes preclude model release, the emergence of large-scale public ECG datasets now makes it feasible to develop competitive, open-weight pretrained models. \\

In this study, we present ECG-FM, a transformer-based ECG foundation model pretrained using a hybrid self-supervised learning method with 1.4 million ECG segments. Our results indicate that ECG-FM is a robust, generalizable, and functionally discriminative model which can alleviate the need for large annotated datasets. By releasing our model and benchmark task—together with strong performance and rigorous evaluation standards—we aim to lower the barrier for ECG foundation model adoption, facilitate benchmarking efforts, and encourage open-weight practices.

\section{Method}

\subsection{Data}
We collected a total of 1.5 million standard 12-lead ECGs from the UHN-ECG, PhysioNet 2021 \supercite{PhysioNet2021paper,physionet2021,physionet}, and MIMIC-IV-ECG \supercite{mimicIVpaper, physionet} datasets. Specifically, we include six datasets in PhysioNet 2021: CPSC, CPSC-Extra, PTB-XL, Georgia, Ningbo, and Chapman. The PTB and St Petersburg INCART datasets are excluded due to having few long samples with inconsistent sampling rates. The MIMIC-IV-ECG v1.0 database contains many 10 s ECGs collected from the Beth Israel Deaconess Medical Center. PhysioNet 2021 and MIMIC-IV-ECG are two public data sources used for our pretraining dataset, while UHN-ECG was utilized only for downstream tasks.

\subsubsection{UHN-ECG}
UHN-ECG is a newly assembled private, institutional dataset containing 622k ECG recordings from 211k patients who were seen in the emergency department and/or admitted to hospital between January 2010 and December 2018. This 9-year dataset was collected at Toronto General Hospital and Toronto Western Hospital, which are two acute care hospitals that have emergency departments, cardiology wards, and coronary care units. These exist as part of the University Health Network (UHN), a network of academic hospitals located in Toronto, Canada. All recordings are 10 s, where 88.8\% have an original sampling frequency of 500 Hz and the remaining have 250 Hz. Every ECG has a cardiologist over-read and there are several associated clinical reports and auxiliary data modalities which make for excellent label availability. \\

Approximately 12.8\% of the ECGs were labeled with poor data quality on their interpretations. This includes various artifacts, muscle interference, as well as lead misplacement and reversal. We noticed that interpretation may be attempted regardless, so we opted to predict through poor data quality during both training and evaluation, so as to retain these samples and produce a model robust to real-world clinical environments.

\subsubsection{ECG preprocessing}
We extracted raw waveforms and tabularized ECG metadata, including sample rates, sample size, and patient demographic information wherever available. We resampled the waveforms at 500 Hz using linear interpolation, performed z-score normalization, and segmented the signals into non-overlapping 5 s segments to produce the model inputs. Segmenting the ECGs is necessary to generate positive pairs for the CMSC contrastive objective, and since UHN-ECG and MIMIC-IV-ECG both contain 10 s recordings, the vast majority of ECGs are segmented in half.

\begin{figure}
    \centering
    \includegraphics[scale=0.55]{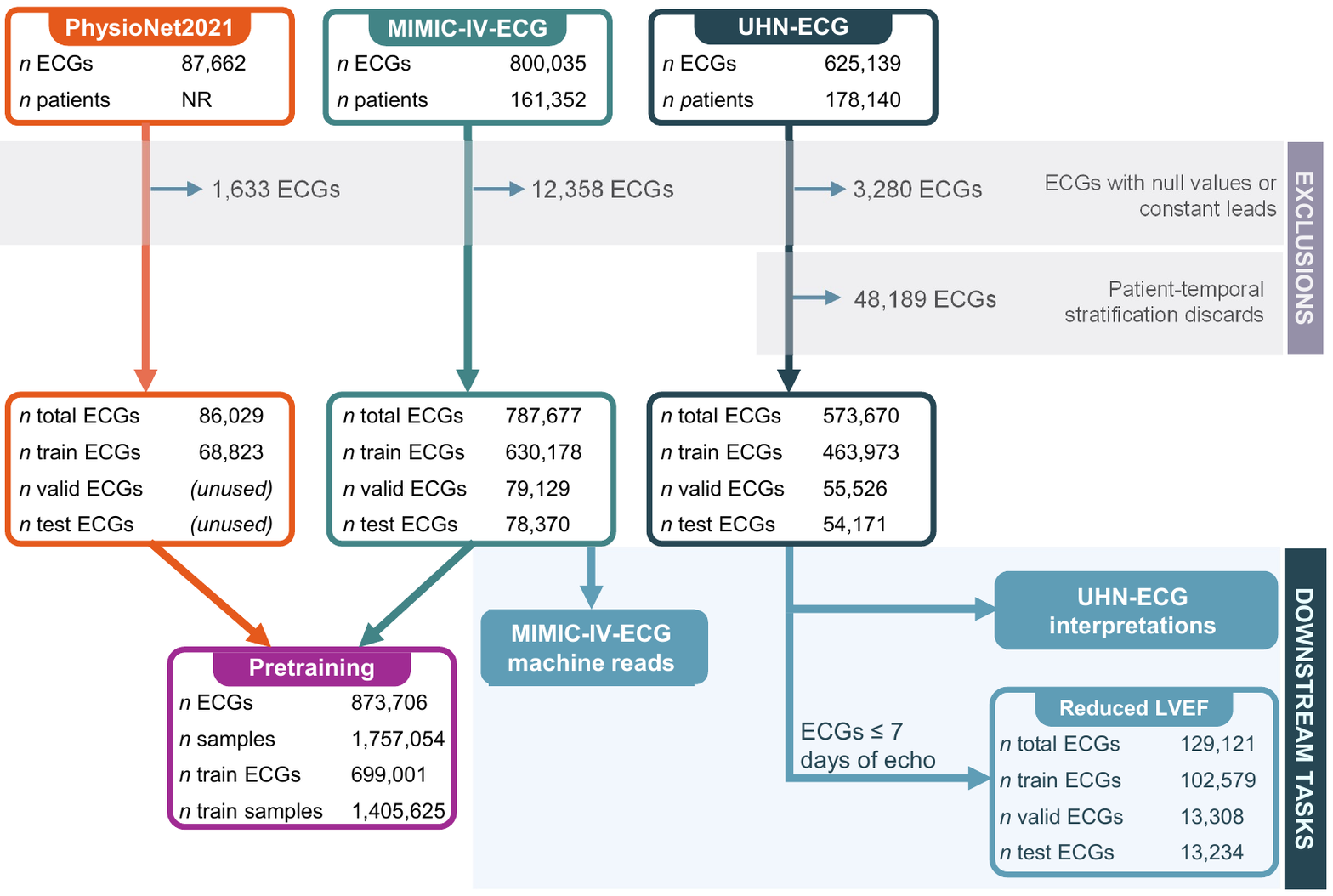}
    \caption{\textbf{Cohort and sample selection.} This flow diagram shows the data sources and ECG exclusion criteria, as well as the dataset partitioning. The pretraining cohort combines samples from public datasets PhysioNet2021 and MIMIC-IV-ECG. Downstream task cohorts utilize MIMIC-IV-ECG and UHN-ECG datasets, where the reduced LVEF task undergoes filtering according to task-specific label availability.}
    \label{fig:cohort}
\end{figure}

\subsection{Cohort curation}
We removed ECGs having null values or constant-valued leads across all three datasets. To maintain a representative distribution of in-hospital populations and avoid introducing selection bias, there were no additional exclusions based on ECG or patient characteristics. \\

Each dataset was stratified to approximately produce 80\%/10\%/10\% splits for training, validation, and testing, respectively. These partitions were respected across all experiments, whether pretraining or finetuning. The method by which each dataset was split was dependent on its available metadata. The UHN-ECG dataset was stratified by patient as well as temporally, ensuring there is no overlap in patients or ECG acquisition periods across splits. Therefore, label leakage is avoided by evaluating on different subjects and our retrospective evaluation more closely resembles a prospective evaluation (see supplementary Figure \ref{fig:temporal-split} for more details). MIMIC-IV-ECG was split only by patient due to having imprecise acquisition dates. PhysioNet 2021 was split at random, where its evaluative sets were not used for the purposes of this work. We did not pretrain on the UHN-ECG dataset so that we may release our pretrained model while respecting patient privacy, as well as to explore the cross-dataset generalizability of ECG-FM.

\subsection{Model architecture}
ECG-FM has 90.9 million parameters and uses the wav2vec 2.0 \supercite{wav2vec2} model architecture, which consists of a multi-layer CNN feature extractor and a BERT-like transformer encoder. The feature extractor effectively embeds portions of the raw signal, generating latent representations $\mathbf{z}_t$ which are fed to a transformer encoder to create local contextualized representations $\mathbf{c}_t$. \\

The feature encoder specifically contains 4 blocks, each composed of a convolutional layer having 256 channels, a stride of 2, and a kernel length of 2, followed by layer normalization and the GELU activation function. Relative positional embeddings are learned using a convolutional layer having 128 filters and 16 groups, which are then added to the latent representations. For the transformer encoder, we selected hyperparameters consistent with the BERT-Base encoder, having 12 transformer block layers, an embedding dimension of 768, 12 self-attention heads, and a feed-forward network dimension of 3,072.

\subsection{Pretraining method}
We build upon the work of \authorcite{wcr}, adopting their pretraining method and engaging in open-source collaboration\footnote{\url{https://github.com/Jwoo5/fairseq-signals/}}. This hybrid self-supervised learning method combines the continuous signal masking objective proposed in wav2vec 2.0 \supercite{wav2vec2}, the contrastive CMSC objective originally introduced by CLOCS \supercite{clocs}, and the RLM augmentation presented by \authorcite{wcr}. This approach, previously named W2V+CMSC+RLM, will henceforth be referred to as WCR for brevity.

\subsubsection{wav2vec 2.0}
Inspired by masked language modeling, wav2vec 2.0 masks spans of CNN latent representations $\mathbf{z}_t$. Each token has a 6.5\% probability of being a starting index, where if selected, we mask the subsequent 10 tokens. This results in approximately $49\%$ of a sample's tokens being masked. Rather than directly using the masked $\mathbf{z}_t$ as the target, we adopt the wav2vec 2.0 quantization module to help remove detailed artifacts which would otherwise make the task easier, thereby hurting generalizability \supercite{wav2vec2}. Specifically, we quantize $\mathbf{z}_t$ to $\mathbf{q}_t$ during pretraining using two trainable codebooks of 320 codes. A codebook diversity loss is applied to encourage more equal usage of codebook entries. In order to capture local context, a contrastive loss is used to enable the distinction of a true latent from a number of distractors. For each masked token, we maximize the cosine similarity between quantized target $\mathbf{q}_t$ and its corresponding local contextualized representation $\mathbf{c}_t$, while minimizing it over distractors $\mathbf{\tilde{q}} \sim \mathbf{Q}_t$, which are sampled from the set of all masked token quantized targets $\mathbf{Q}_t$. This requires the model to learn how to leverage local contextual information to infer $\mathbf{c}_t$ based on nearby unmasked representations.

\subsubsection{CMSC} CMSC is a contrastive objective applied between global representations, treating temporally adjacent ECG segments as positive pairs and deriving negative pairs from other segments \supercite{clocs}. As this unique strategy does not involve augmentation, it circumvents the issue of faulty alignment. Further, by encouraging consistent representations between consecutive ECG segments, CMSC is designed to promote temporal invariance and prioritize the capture of functionally relevant information over superficial presentation.

\subsubsection{RLM} \authorcite{wcr} introduced Random Lead Masking (RLM), an ECG-specific augmentation wherein each individual lead is masked at random with probability $p = 0.5$. Though not explored in this study, \authorcite{wcr} demonstrated how the RLM augmentation ameliorates finetuning on arbitrary lead subsets, suggesting that ECG-FM is applicable not only with standard 12-lead ECGs, but also in contexts where only a reduced lead set is available. \\

\begin{figure}
    \centering
    \includegraphics[scale=0.65]{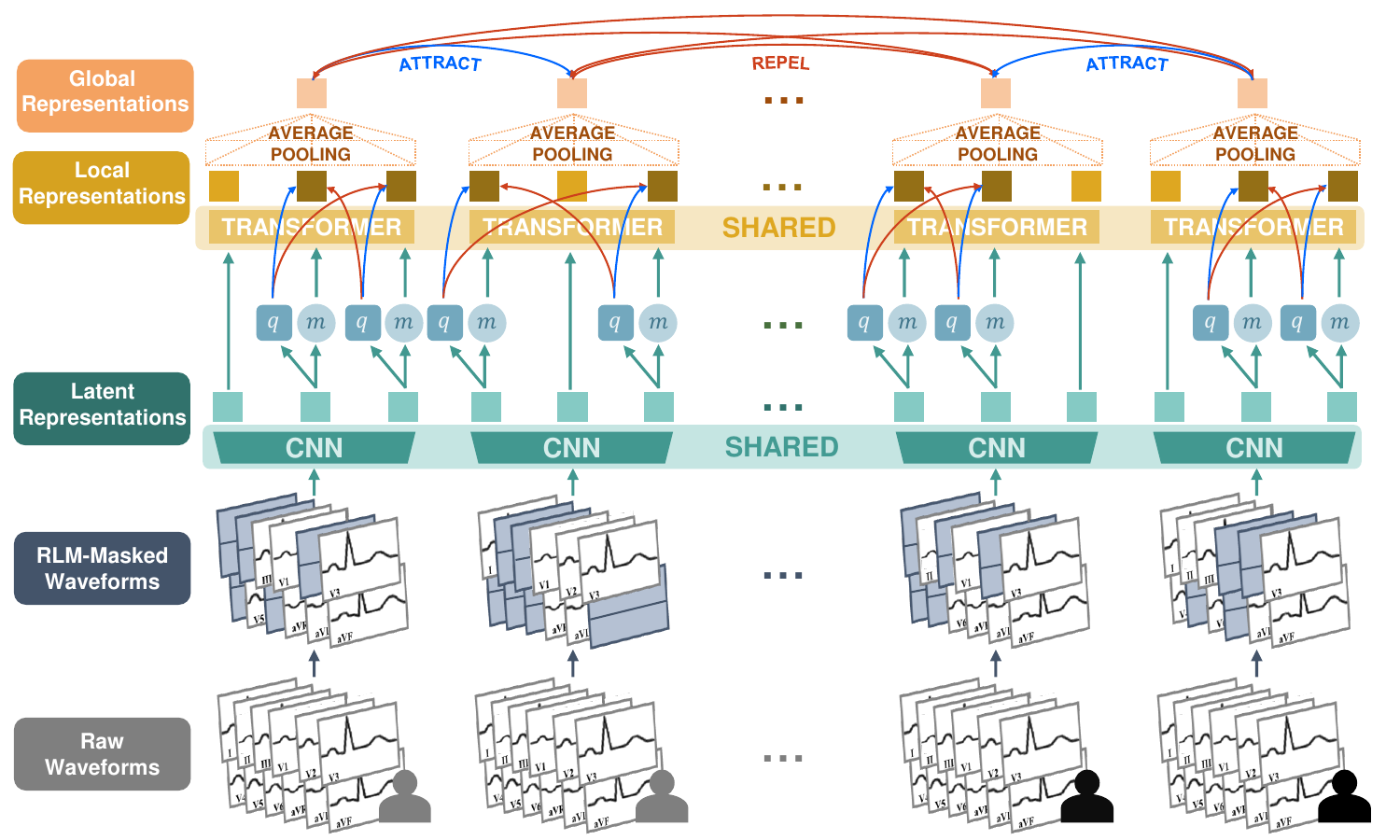}
    \caption{\textbf{Framework illustration.} Raw waveforms are inputted and individual leads are randomly masked. A convolutional feature encoder generates latent representations that feed into a transformer encoder, producing local representations that are then average-pooled to create global representations. Latent representations are randomly masked in spans as $m$, and are quantized to $q$. We then apply a local contrastive loss attracting each $q$ to its corresponding local representation, using a subset of other $q$ as the negative samples, or distractors. A batch of four ECG inputs, making up two positive pairs of temporally-adjacent ECG segments, are shown to visualize the CMSC global contrastive loss acting on the global representations across samples. Positive and negative contrastive learning pair relationships are depicted using blue and red arrows, respectively.}
    \label{fig:architecture}
\end{figure}

Pretraining ECG-FM using WCR is driven by several motivations. The intra-ECG wav2vec 2.0 contrastive loss is employed with the intention to enhance the capture of local patterns. The inter-ECG CMSC contrastive loss is meant to foster more functionally discriminative global representations, additionally promoting temporal invariance. RLM is incorporated as an augmentation strategy with the aim of enhancing model robustness and allowing for finetuning on reduced lead sets \supercite{wcr}.

\subsection{Downstream tasks}
To demonstrate that ECG-FM is useful for a variety of downstream applications, we selected several classification tasks for our evaluation.

\subsubsection{UHN-ECG interpretation}

As primary readers, cardiologists demonstrate greater interpretation accuracy and less interobserver variability in comparison to physicians with less specialized training \supercite{RN938,cnnlvef35}. In the UHN-ECG dataset, each ECG is accompanied by a cardiologist's interpretation, provided as an over-read of an automated analysis. We extracted binary labels from these free-text expert interpretations. As the interpretations are well-structured, we developed a text-parsing system based on pattern matching and a knowledge graph to produce more accurate labels. Further details can be found in supplementary Section \ref{sec:interpretation-parsing}.


\subsubsection{MIMIC-IV-ECG machine reads}
MIMIC-IV-ECG utilizes machine measurements to automatically generate ECG reports. For this task, we generate our labels from these reports using the same text-parsing system as with the UHN-ECG interpretation task, adjusting the patterns to accommodate dataset-specific terminology. We focused on all the same labels as in the UHN-ECG interpretation task, given they were available and had a fairly comparable outcome prevalence. Our meticulous conversion of free-text interpretations into binary labels culminates in an accessible benchmark task.

\subsubsection{UHN-ECG reduced LVEF}
Heart failure is a major contributor of morbidity and mortality worldwide, with approximately 50\% of cases being heart failure with reduced ejection fraction (HFrEF) \supercite{lvef-Murphy2020}. The heightened risk of mortality associated with HFrEF, even in asymptomatic cases, underscores the significance of early detection for timely intervention \supercite{lvef-yao,lvef-yao-eagle}. We predict low, or reduced, LVEF as a step towards early screening for left ventricular systolic dysfunction and HFrEF. Using regex to extract LVEF percentages from echocardiography reports, we generated labels at various common LVEF thresholds: \( \leq 50\%\), \( \leq 40\%\), \( \leq 35\%\), and \( \leq 30\%\). We paired each ECG with its closest associated echocardiography report which indicates an LVEF percentage, taking only those ECG samples with a valid report within $\pm$7 days of acquisition.

\subsection{Experiments}
The ECG-FM model was pretrained on 1,405,625 samples using three A100 80GB GPUs applying distributed data parallelism. The 1026 batch size was composed of 171 positive pairs, each consisting of two segments, distributed across the three GPUs. A fixed learning rate schedule was used, where it was initially set at $1 \times 10^{-4}$ for the first 5 epochs, reduced to $8 \times 10^{-5}$ from epoch 6 to 200, and further decreased to $5 \times 10^{-5}$ for epochs 201 to 240. Training concluded after 240 epochs, totaling a computational wall time of 9.51 days. \\

We ran several suites of ECG-FM multi-label classification experiments for each downstream task. Loss balancing was applied using weights inversely proportional to the outcome prevalence. We initialized our \textit{Full} models with the pretrained weights and performed full finetuning, wherein all model weights are updated, using a learning rate of $1\times10^{-6}$. For \textit{Random Init.}, we randomly initialized the models and then performed full finetuning with a learning rate of $1\times10^{-5}$. To maintain a fair comparison, the weight initialization and learning rates were the only experimental differences between the \textit{Full} and \textit{Random Init.} experiments. In the \textit{Linear} experiment, we performed linear probing with learning rate $1\times10^{-5}$. For this frozen evaluation, pretrained model embeddings are extracted and fed as inputs to a single linear layer to generate predictions. \\

We also employed two task-specific, ResNet-based baseline models, which were highly competitive in the PhysioNet Challenge 2021 \supercite{physionet2021}, to contrast ECG-FM performance at various data scales. The \textit{Nejedly} baseline is a ResNet with multi-head attention which follows the configuration seen in \authorcite{Nejedly2022} and is trained with learning rate $1\times10^{-4}$. The \textit{SE-WRN} baseline implementation is similar to that of \authorcite{SE-WRN} and also used a learning rate $1\times10^{-4}$. \\

Data scaling experiments were performed by taking a percentage subset of ECGs in the training set while maintaining the same evaluative sets. We ran experiments on 50\%, 10\%, and 1\% of the training set ECGs for the interpretation tasks, as well as 50\% and 10\% for the reduced LVEF task. \\

All downstream tasks ran on a single A100 80GB GPU using a batch size of 256. Experiments used the Adam optimizer \supercite{kingma2017adam} with $\beta_1=0.9$, $\beta_2=0.98$. Checkpoints were selected according to which had the best AUPRC on the validation set. Thresholds were computed to achieve target recall values on the validation set. Aside from the aggregated results shown in supplementary Table \ref{tab:aggregation}, all evaluative metrics are computed by randomly sampling a single segment per ECG. Considering certain highly imbalanced labels, we opted to report label-averaged AUPRC for most summary metrics.

\section{Results}

\begin{figure}
    \centering
    \includegraphics[scale=0.5]{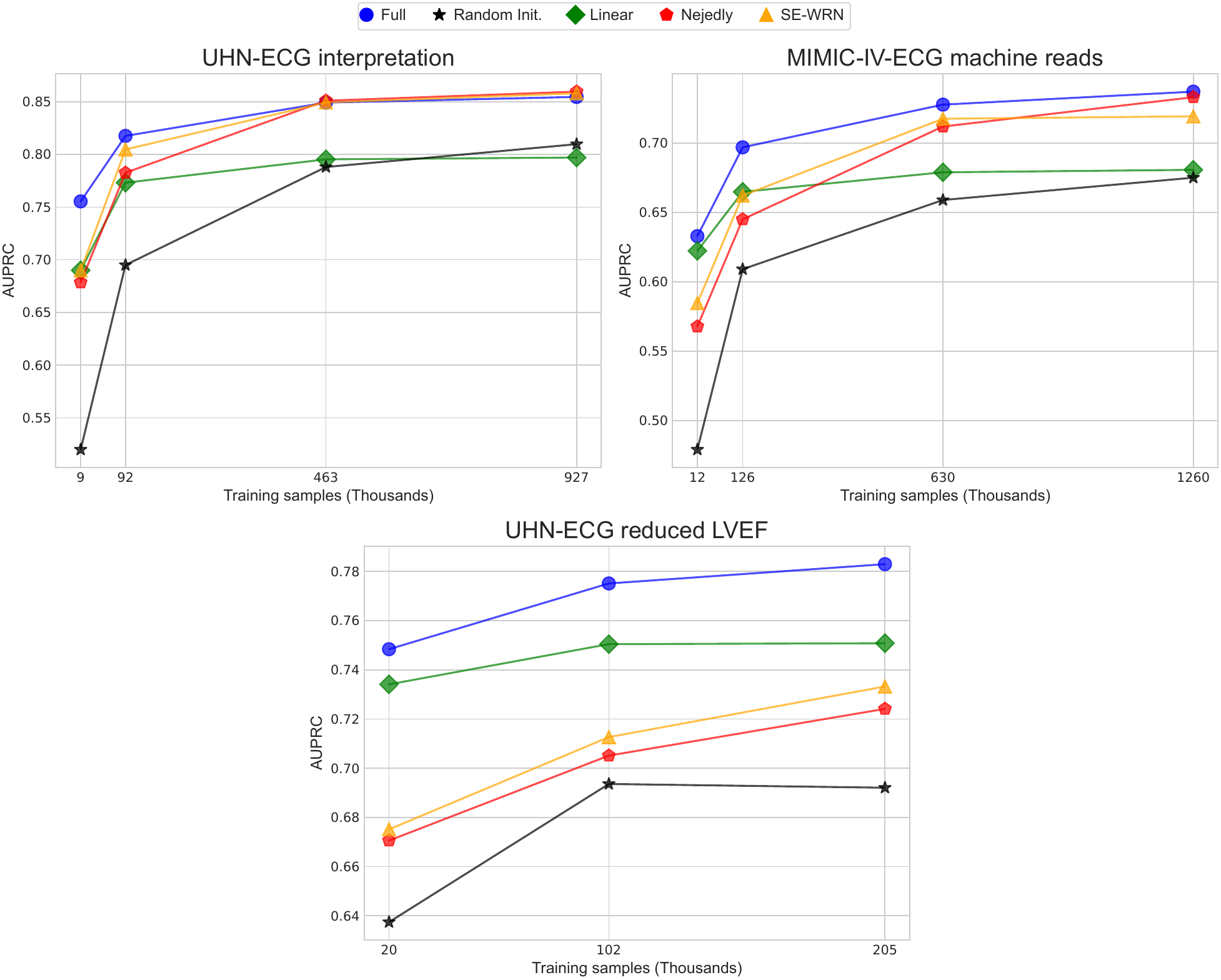}
    \caption{\textbf{Data scaling results.} Label-averaged AUPRC across experiment suites and training dataset sizes for all three tasks.}
    \label{fig:scaling}
\end{figure}

\subsection{Data scaling}

\subsubsection{Pretraining benefits}
The \textit{Linear} results demonstrate that our pretrained model embeddings encode rich, task-relevant information. The \textit{Random Init.} performance is relatively poor with few training samples, confirming that ECG-FM's pretraining is responsible for its superior data efficiency. At the smallest training set sizes, \textit{Linear} outperforms the baselines and performs comparably to \textit{Full} in the MIMIC-IV-ECG machine reads and UHN-ECG reduced LVEF tasks; however, its performance plateaus because it lacks the representational capacity necessary to exploit additional downstream data. Though task-dependent, this indicates that the benefits of WCR pretraining are quite significant in smaller data regimes and generalize well to new datasets.

\subsubsection{Downstream performance}
Across all tasks, \textit{Full} outperforms the baselines given approximately 100,000 training samples. However, for the UHN-ECG interpretation and MIMIC-IV-ECG machine reads tasks, the \textit{Full} and baseline model performances trend to a similar plateau. We conclude that the ECG-FM model can confidently outperform strong task-specific models on downstream tasks in the small-to-medium-scale data regime, but may not provide significant value downstream given sufficiently large task-specific datasets as released.

\begin{figure}
    \centering
    \includegraphics[scale=0.58]{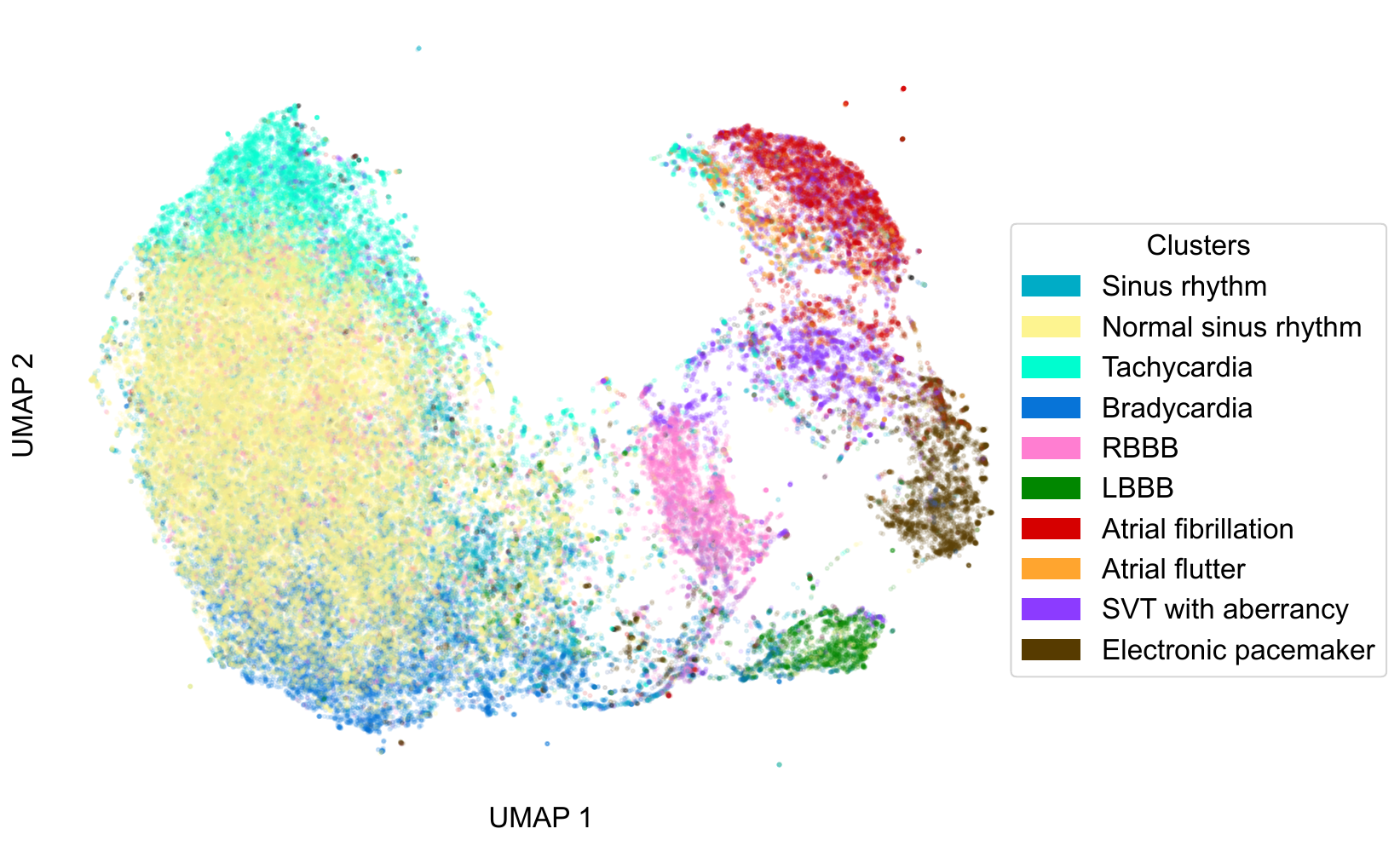}
    \caption{\textbf{Pretrained latent space UMAP.} A UMAP visualization of pretrained ECG-FM global representations from ECGs in the UHN-ECG dataset. An additive color scheme employs select labels from the UHN-ECG interpretation task to enable a latent space analysis, wherein some prioritization was performed to prevent label overlap from reducing readability.}
    \label{fig:embs}
\end{figure}

\subsection{Latent space analysis} \label{sec:latent_space_analysis}
The UMAP visualization of the UHN-ECG dataset in Figure \ref{fig:embs} highlights ECG-FM's ability to encode relationships in rhythm, heart rate, and pathology for an unseen dataset. This latent space depicts our pretrained model—which has never been trained on any labels—making the label overlay strictly evaluative. \\

A distinct \textit{Normal sinus rhythm} cluster sits within a broader \textit{Sinus rhythm} region, with \textit{Bradycardia} and \textit{Tachycardia} samples forming its extremities and encoding heart rate along the vertical axis. Non-sinus tachycardias like \textit{Atrial fibrillation} and \textit{Atrial flutter} occupy a separate upper region, while conduction defects, such as the bundle branch blocks, cluster more centrally. This spatial arrangement, along with the close proximity of related tachycardias, underscores ECG-FM's ability to capture functional similarity, and possibly co-occurrence, as it recognizes and differentiates pathological rhythms and conditions. Although \textit{Electronic pacemaker}'s strong clustering might suggest a decreased recognition of other positive labels in these samples, an evaluative pacemaker-only subset revealed no performance drop for other labels, indicating a robust encoding of diverse relationships beyond those visually evident in Figure \ref{fig:embs}. Overall, these observations highlight ECG-FM's capacity to produce physiologically meaningful and functionally discriminative representations as a direct result of WCR pretraining.

\subsection{Downstream tasks}
\subsubsection{UHN-ECG interpretation}
As evidenced in supplementary Table \ref{tab:ecg_interpretation}, we achieve strong performance across numerous labels using a cohort representative of real in-hospital populations. Our results demonstrate that ECG-FM is resilient even to low-quality recordings, with a considerable 12.4\% of the test set labeled with \textit{Poor data quality}. Our findings suggest that performance is influenced by a condition's outcome prevalence, the subtlety of its ECG signature, and the complexity of its diagnostic criteria. \\

\subsubsection{MIMIC-IV-ECG machine reads}
To evaluate the quality of this task for the purposes of benchmarking, it is helpful to compare its results in supplementary Table \ref{tab:mimic_iv_ecg_machine_reads} to those labels common to the UHN-ECG interpretation task. Across most labels, performance is comparatively lower on this task, with the UHN-ECG interpretations showing notably stronger results for \textit{Poor data quality}, \textit{Atrial flutter}, and \textit{Ventricular tachycardia}. This disparity likely stems from differences in label quality, given the reliance on automated machine measurements. For \textit{Bifascicular block}, however, this task achieves a higher AUPRC, and several other labels see comparable performance. This indicates that, despite using machine reads, the resulting labels rely on patterns which are consistently recognizable by ECG-FM, indicating that this is an entirely practical benchmarking task.

\subsubsection{UHN-ECG reduced LVEF}
Supplementary Table \ref{tab:uhn_lvef} presents performances across all experiment suites, where our \textit{Full} experiment outperforms both baselines across all data scales and labels. It is unclear whether the performance gap between the baselines and \textit{Full} would similarly close, as with the other tasks, given greater task-specific data availability. We see an upwards trend in the \textit{Full} experiment performance which suggests that WCR pretraining would continue to provide benefit given a larger dataset.

\begin{figure}
    \centering
    \includegraphics[scale=0.44]{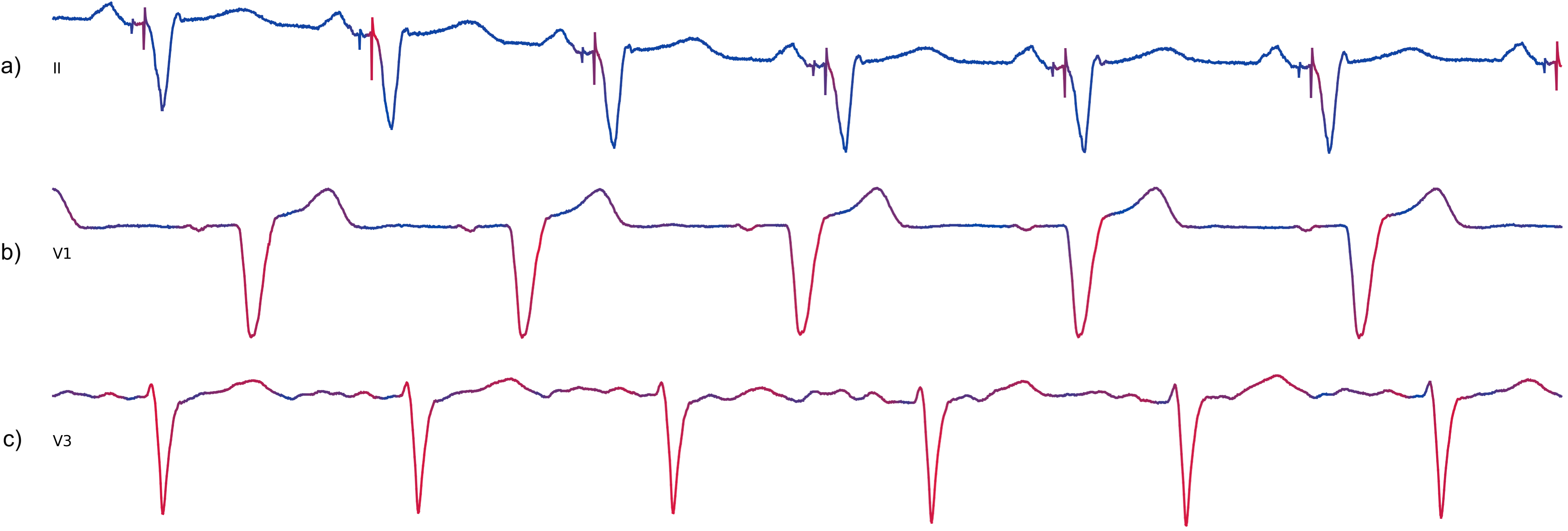}
    \caption{\textbf{Saliency maps.} Distinct 5 s ECG segments colored using corresponding self-attention weight activations derived from pretrained, full-finetuned ECG-FM models. Red represents a higher relative activation. (a) UHN-ECG interpretation model activations for an ECG labeled with ventricular pacing (lead II); (b) UHN-ECG interpretation model activations for an ECG labeled with LBBB (lead V1); (c) UHN-ECG reduced LVEF model activations for an ECG labeled with LVEF$\leq$30\% (lead V3).}
    \label{fig:saliency}
\end{figure}

\subsection{Saliency maps}
We generated attention-based saliency maps by extracting attention weights from the final transformer encoder self-attention layer, averaging across attention heads, and projecting these into the input space. The visualizations in Figure \ref{fig:saliency} serve as a heuristic for relative input importance. \\

We observed that our models attend to consistent regions across cardiac cycles. For instance, in predicting \textit{Ventricular pacing}, attention is often placed on the pacing spikes—a hallmark feature of paced rhythms—suggesting a sensitivity to clinically relevant localized features. While saliency maps alone cannot fully explain model decision-making, these displays of preferential attention imply that our model makes effective use of local contextual information in its prediction.

\subsection{Independent evaluation}
\authorcite{meng2025fusion} performed an independent preprint study on the private MEDIC prehospital ambulance dataset, comprising 5,813 10 s 12-lead ECGs with a 20\% prevalence of acute coronary syndrome (ACS). In a frozen evaluation, they extracted pretrained model embeddings and fed them to an XGBoost classifier for ACS detection. The pretrained ECG-FM model was contrasted with that of ST-MEM (Spatio-Temporal Masked Electrocardiogram Modeling), a generative-only SSL method \supercite{st-mem}. ECG-FM achieved an AUPRC of $0.667\pm0.024$, surpassing ST-MEM's $0.594\pm0.013$ \supercite{meng2025fusion}. \\

\section{Discussion}
We demonstrate strong performance of ECG-FM on clinically relevant tasks. The UHN-ECG interpretation experiment marks a step toward full, expert-level 12-lead interpretation, while the reduced LVEF task highlights ECG-FM's potential to inform the rapid development of medical management plans. Evaluated in a simulated prospective setting that reflects representative in-hospital populations and tracing quality, these UHN-ECG tasks validate ECG-FM's cross-dataset generalizability. The independent results from \authorcite{meng2025fusion} suggest that ECG-FM is discriminative for early ACS triage in the noisier prehospital setting. We release models, manifests, and labeling code surrounding the MIMIC-IV-ECG machine reads task so that it may serve as a benchmark task. \\

Our data scaling experiments demonstrate a rapid adaptability to downstream tasks and reduced reliance on labeled data. ECG-FM consistently outperforms conventional task-specific methods in the small-to-medium-scale data regime, underscoring the benefit of WCR pretraining. The linear probing experiments confirm that the pretrained embeddings encode task-relevant information. Together with the independent ACS findings \supercite{meng2025fusion}, these results suggest that ECG-FM can act as a robust, competitive feature-set generator for diverse clinical applications. \\

Our results suggest that ECG-FM commands local and global contextual information effectively. In our saliency maps, it displays relevant, preferential attention to the same regions across the cardiac cycle. Our latent space analysis exhibits pretrained embeddings which are functionally discriminative and indicative of ECG-FM's ability to capture underlying cardiac function. Such explorations form a basis to demystify internal model workings and improve model interpretability. \\

The CMSC objective addresses the issue of faulty alignment and promotes temporal invariance; however, its positive pair strategy of adopting consecutive segments necessitates ECG-FM accept 5 s ECG inputs, as our datasets largely contain 10 s recordings. Since our labels are not segment-specific, this necessarily impacts our experimental setup by creating a mismatch between the input and label. In turn, this can underrate our model's capabilities, which we explore in supplementary Section \ref{sec:aggregation} by aggregating same-sample segment logits. This aggregation greatly improves performance on select labels, such as a 16.65\% increase in AUPRC when predicting \textit{Premature ventricular contraction} (PVC), showcasing that the effects of this limitation are label-specific and seemingly mitigable through simple techniques. \\

We envision many promising avenues for extending this work. As an ECG encoder, ECG-FM is well-suited for integration into multimodal foundation models that leverage complementary data sources, such as medical imaging or clinical notes. In particular, pairing ECG-FM with a multimodal large language model tailored for ECG analysis could enable automated generation of ECG interpretations and other textual reports, enhancing utility and interpretability through greater diagnostic precision, improved uncertainty quantification, and potentially question-answering capabilities.

\subsubsection{Data availability} PhysioNet 2021 v1.0.3 is available for public download (\url{https://doi.org/10.13026/34va-7q14}), as is MIMIC-IV-ECG v1.0 (\url{https://doi.org/10.13026/4nqg-sb35}). The UHN-ECG dataset is not available for public use. Model weights are available on our GitHub for our pretrained model and downstream MIMIC-IV-ECG machine reads models. Downstream UHN-ECG models cannot be made available due to privacy concerns. Code for data preprocessing, model training, model inference, and experiment reproduction is available, as are tutorial notebooks. Refer to \url{https://github.com/bowang-lab/ECG-FM/}.

\subsubsection{Contributions} K.M conceived the study, performed all experiments, analyzed the data, and drafted the manuscript. S.M contributed to clinical aspects of the study design and aided in result interpretation and communication. A.T advised the machine learning methodology, interpreted model outputs, and aided in methodology communication. B.R contributed to study conception and clinical interpretation of results. B.W advised the machine learning methodology and interpreted model outputs. All authors critically reviewed and revised the manuscript.

\subsubsection{Competing interests} The authors declare no competing interests.

\newpage
\printbibliography

\newpage
\section{Supplementary Material}

\begin{table}
\caption{\textbf{Abbreviations.}} \label{tab:abbrev}
\centering
\begin{tabular}{l|l}
\hline
ECG  &	Electrocardiogram \\
UHN  &	University Health Network \\
WCR  &	W2V+CMSC+RLM \\
W2V  &	wav2vec 2.0 \\
CLOCS & Contrastive Learning Of Cardiac Signals \\
CMSC &  Contrastive Multi-Segment Coding \\
RLM  &	Random lead masking \\
HFrEF & Heart failure with reduced ejection fraction \\
LVEF &	Left ventricular ejection fraction \\
PVC  &	Premature ventricular contraction \\
SVT  &  Supraventricular tachycardia \\
AP   &	Accessory pathway \\
AV   &	Atrioventricular \\
RBBB &	Right bundle branch block \\
LBBB &	Left bundle branch block \\
\hline
\end{tabular}
\end{table}

\begin{figure}
    \centering
    \includegraphics[scale=0.55]{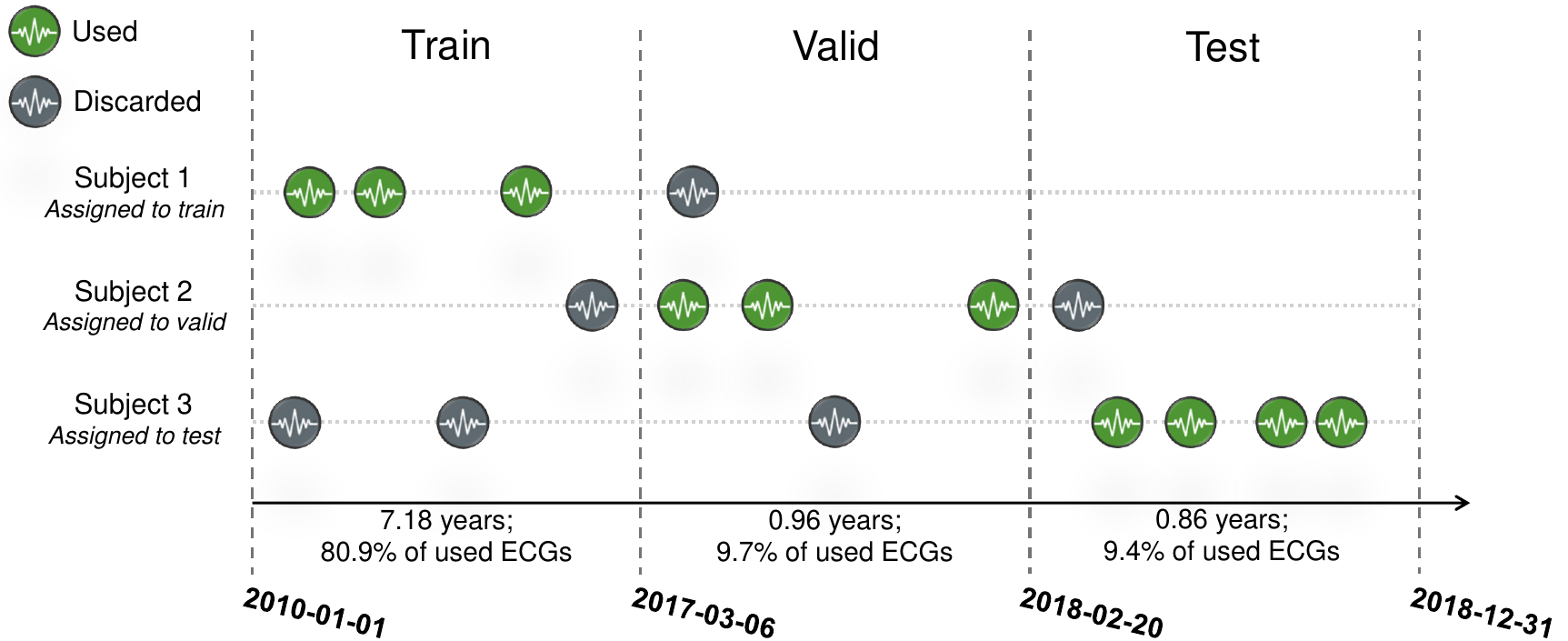}
    \caption{\textbf{Illustration of UHN-ECG patient-temporal splits.} UHN-ECG dataset splits have no overlap temporally, where temporal cutoffs were estimated to generate an approximate 80/10/10 split. Subjects were assigned to whichever split contained the majority of their ECGs. Patient overlap is not permitted between splits to prevent any possibility of label leakage.}
    \label{fig:temporal-split}
\end{figure}

\begin{table}
\caption{\textbf{Outcome prevalence.} Positive label frequency rates (\%), as separated by split and task.} \label{tab:prevalence}
\centering
\begin{tabular}{l|ccc}
\hline
\textbf{Label}       & \textbf{Train} & \textbf{Validation} & \textbf{Test} \\
\hline
\textbf{UHN-ECG interpretations} &  &  &  \\
\hline
\textbf{Poor data quality} & 0.126 & 0.151 & 0.124 \\
\textbf{Sinus rhythm} & 0.814 & 0.817 & 0.823 \\
\textbf{Normal sinus rhythm} & 0.465 & 0.467 & 0.472 \\
\textbf{PVC} & 0.064 & 0.070 & 0.066 \\
\textbf{Tachycardia} & 0.218 & 0.220 & 0.214 \\
\textbf{Ventricular tachycardia} & 0.001 & 0.001 & 0.001 \\
\textbf{SVT with aberrancy} & 0.055 & 0.057 & 0.049 \\
\textbf{Atrial fibrillation} & 0.094 & 0.089 & 0.084 \\
\textbf{Atrial flutter} & 0.022 & 0.024 & 0.022 \\
\textbf{Bradycardia} & 0.120 & 0.120 & 0.124 \\
\textbf{Accessory pathway conduction} & 0.115 & 0.112 & 0.106 \\
\textbf{AV block} & 0.093 & 0.095 & 0.096 \\
\textbf{1st degree AV block} & 0.066 & 0.066 & 0.070 \\
\textbf{2nd degree AV block} & 0.024 & 0.025 & 0.023 \\
\textbf{Bifascicular block} & 0.018 & 0.016 & 0.019 \\
\textbf{RBBB} & 0.104 & 0.107 & 0.108 \\
\textbf{LBBB} & 0.038 & 0.039 & 0.034 \\
\textbf{Myocardial infarction} & 0.193 & 0.187 & 0.199 \\
\textbf{Electronic pacemaker} & 0.067 & 0.063 & 0.062 \\
\textbf{Ventricular pacing} & 0.047 & 0.048 & 0.046 \\
\textbf{Atrial pacing} & 0.020 & 0.015 & 0.016 \\
\hline
\textbf{MIMIC-IV-ECG machine reads} &  &  &  \\
\hline
\textbf{Poor data quality} & 0.026 & 0.027 & 0.026 \\
\textbf{Sinus rhythm} & 0.812 & 0.811 & 0.809 \\
\textbf{PVC} & 0.065 & 0.065 & 0.067 \\
\textbf{Tachycardia} & 0.209 & 0.210 & 0.210 \\
\textbf{Ventricular tachycardia} & 0.001 & 0.001 & 0.001 \\
\textbf{SVT with aberrancy} & 0.042 & 0.043 & 0.040 \\
\textbf{Atrial fibrillation} & 0.101 & 0.103 & 0.103 \\
\textbf{Atrial flutter} & 0.018 & 0.019 & 0.018 \\
\textbf{Bradycardia} & 0.130 & 0.132 & 0.132 \\
\textbf{Accessory pathway conduction} & 0.119 & 0.123 & 0.120 \\
\textbf{AV block} & 0.078 & 0.085 & 0.081 \\
\textbf{1st degree AV block} & 0.073 & 0.079 & 0.076 \\
\textbf{Bifascicular block} & 0.030 & 0.033 & 0.030 \\
\textbf{RBBB} & 0.082 & 0.088 & 0.081 \\
\textbf{LBBB} & 0.037 & 0.039 & 0.038 \\
\textbf{Myocardial infarction} & 0.224 & 0.226 & 0.225 \\
\textbf{Electronic pacemaker} & 0.039 & 0.036 & 0.042 \\
\hline
\textbf{UHN-ECG reduced LVEF} &  &  &  \\
\hline
\textbf{LVEF$\boldsymbol\leq$30\%} & 0.245 & 0.141 & 0.156 \\
\textbf{LVEF$\boldsymbol\leq$35\%} & 0.262 & 0.202 & 0.214 \\
\textbf{LVEF$\boldsymbol\leq$40\%} & 0.345 & 0.273 & 0.263 \\
\textbf{LVEF$\boldsymbol\leq$50\%} & 0.488 & 0.420 & 0.394 \\
\hline
\end{tabular}
\end{table}

\begin{figure}
    \centering
    \includegraphics[scale=0.28]{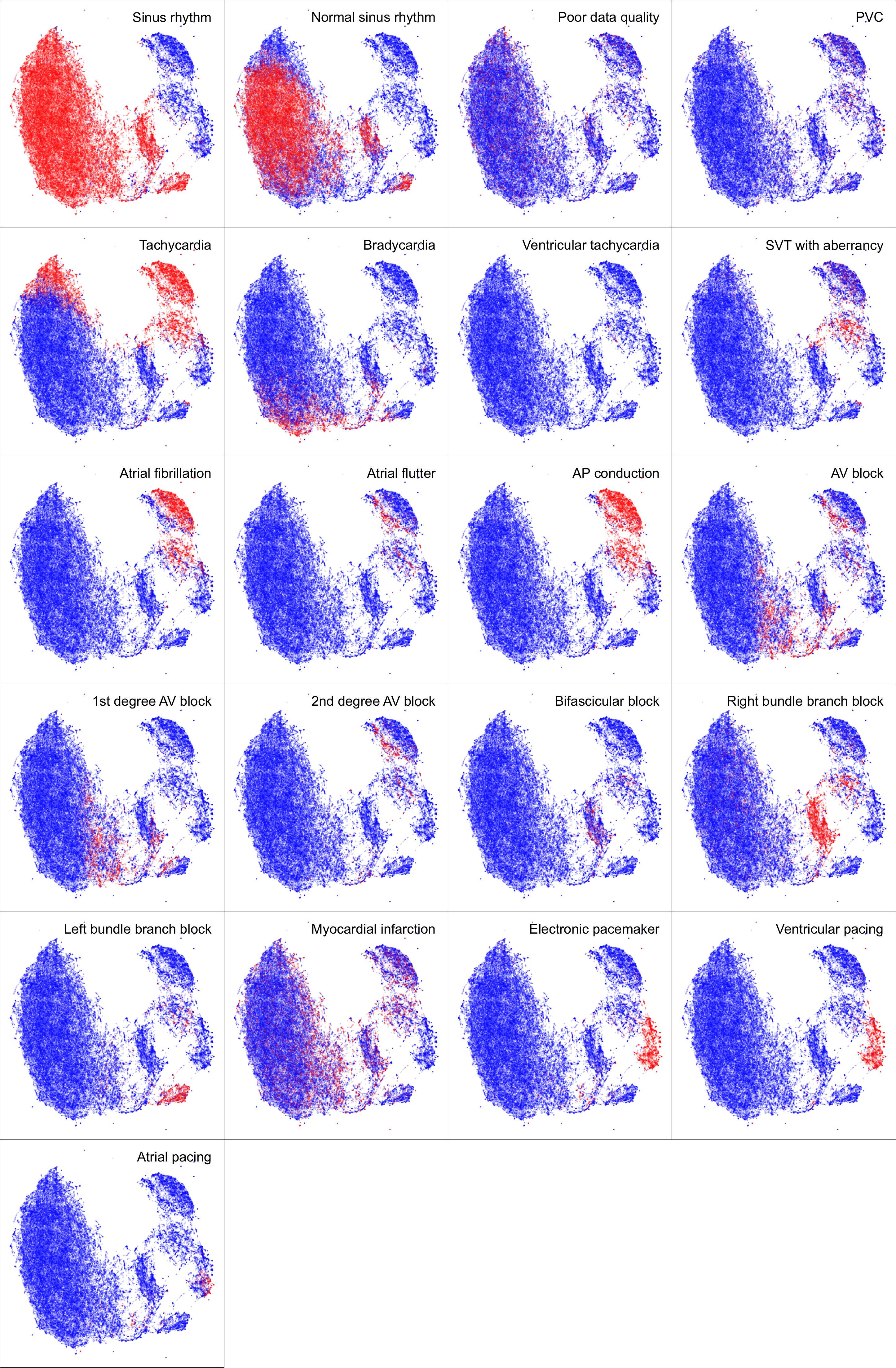}
    \caption{\textbf{Label-specific pretrained latent space UMAPs.} Label-specific UMAP visualizations of pretrained ECG-FM global representations from ECGs in the UHN-ECG dataset. Each subplot shows a different label from the UHN-ECG interpretation task, where positive samples are indicated in red.}
    \label{fig:embs_all_labels}
\end{figure}

\begin{table}
\caption{\textbf{UHN-ECG interpretation test results for \textit{Full}.} Label-specific test performances on our pretrained, full-finetuned model using the full-scale training dataset.} \label{tab:ecg_interpretation}
\centering
\begin{tabular}{lcccccccc}
\hline
\textbf{Label} & \textbf{AUROC} & \textbf{AUPRC} & \textbf{Recall} & \textbf{Precision} & \textbf{F1} & \textbf{Specificity} & \textbf{NPV} & \textbf{Accuracy} \\
\hline
\textbf{Sinus rhythm} & 0.992 & 0.998 & 0.990 & 0.979 & 0.984 & 0.899 & 0.948 & 0.974 \\
\textbf{Normal sinus rhythm} & 0.978 & 0.970 & 0.943 & 0.922 & 0.932 & 0.928 & 0.948 & 0.935 \\
\textbf{Poor data quality} & 0.900 & 0.658 & 0.704 & 0.529 & 0.604 & 0.911 & 0.956 & 0.886 \\
\textbf{PVC} & 0.944 & 0.752 & 0.803 & 0.486 & 0.605 & 0.940 & 0.986 & 0.931 \\
\textbf{Tachycardia} & 0.996 & 0.984 & 0.968 & 0.912 & 0.939 & 0.975 & 0.991 & 0.973 \\
\textbf{Ventricular tachycardia} & 0.992 & 0.453 & 0.755 & 0.356 & 0.484 & 0.999 & 1.000 & 0.999 \\
\textbf{SVT with aberrancy} & 0.983 & 0.807 & 0.850 & 0.575 & 0.686 & 0.968 & 0.992 & 0.962 \\
\textbf{Atrial fibrillation} & 0.996 & 0.968 & 0.957 & 0.814 & 0.880 & 0.980 & 0.996 & 0.978 \\
\textbf{Atrial flutter} & 0.986 & 0.761 & 0.846 & 0.541 & 0.659 & 0.984 & 0.997 & 0.981 \\
\textbf{Bradycardia} & 0.995 & 0.970 & 0.935 & 0.889 & 0.911 & 0.983 & 0.991 & 0.977 \\
\textbf{Accessory pathway conduction} & 0.995 & 0.973 & 0.930 & 0.918 & 0.924 & 0.990 & 0.992 & 0.984 \\
\textbf{AV block} & 0.987 & 0.908 & 0.866 & 0.802 & 0.833 & 0.978 & 0.986 & 0.967 \\
\textbf{1st degree AV block} & 0.992 & 0.920 & 0.856 & 0.859 & 0.857 & 0.990 & 0.989 & 0.980 \\
\textbf{2nd degree AV block} & 0.985 & 0.719 & 0.727 & 0.641 & 0.681 & 0.991 & 0.994 & 0.985 \\
\textbf{Bifascicular block} & 0.990 & 0.688 & 0.678 & 0.616 & 0.645 & 0.992 & 0.994 & 0.986 \\
\textbf{RBBB} & 0.991 & 0.939 & 0.907 & 0.814 & 0.858 & 0.975 & 0.989 & 0.968 \\
\textbf{LBBB} & 0.993 & 0.864 & 0.807 & 0.731 & 0.767 & 0.990 & 0.993 & 0.983 \\
\textbf{Myocardial infarction} & 0.939 & 0.820 & 0.810 & 0.667 & 0.732 & 0.900 & 0.950 & 0.882 \\
\textbf{Electronic pacemaker} & 0.983 & 0.931 & 0.969 & 0.235 & 0.379 & 0.799 & 0.998 & 0.810 \\
\textbf{Ventricular pacing} & 0.997 & 0.969 & 0.947 & 0.948 & 0.947 & 0.998 & 0.998 & 0.995 \\
\textbf{Atrial pacing} & 0.996 & 0.920 & 0.939 & 0.788 & 0.857 & 0.996 & 0.999 & 0.995 \\
\hline
\end{tabular}
\end{table}

\begin{table}
\caption{\textbf{MIMIC-IV-ECG machine read test results for \textit{Full}.} Label-specific test performances on our pretrained, full-finetuned model using the full-scale training dataset.} \label{tab:mimic_iv_ecg_machine_reads}
\centering
\begin{tabular}{lcccccccc}
\hline
\textbf{Label} & \textbf{AUROC} & \textbf{AUPRC} & \textbf{Recall} & \textbf{Precision} & \textbf{F1} & \textbf{Specificity} & \textbf{NPV} & \textbf{Accuracy} \\
\hline
\textbf{Poor data quality} & 0.783 & 0.214 & 0.396 & 0.176 & 0.243 & 0.951 & 0.984 & 0.937 \\
\textbf{Sinus rhythm} & 0.968 & 0.991 & 0.959 & 0.967 & 0.963 & 0.863 & 0.832 & 0.940 \\
\textbf{PVC} & 0.908 & 0.610 & 0.721 & 0.423 & 0.533 & 0.929 & 0.979 & 0.915 \\
\textbf{Tachycardia} & 0.968 & 0.910 & 0.903 & 0.858 & 0.880 & 0.960 & 0.974 & 0.948 \\
\textbf{Ventricular tachycardia} & 0.948 & 0.186 & 0.414 & 0.144 & 0.214 & 0.998 & 0.999 & 0.997 \\
\textbf{SVT with aberrancy} & 0.961 & 0.673 & 0.685 & 0.613 & 0.647 & 0.982 & 0.987 & 0.970 \\
\textbf{Atrial fibrillation} & 0.976 & 0.883 & 0.897 & 0.731 & 0.805 & 0.962 & 0.988 & 0.955 \\
\textbf{Atrial flutter} & 0.931 & 0.471 & 0.580 & 0.428 & 0.493 & 0.986 & 0.992 & 0.978 \\
\textbf{Bradycardia} & 0.965 & 0.860 & 0.848 & 0.819 & 0.833 & 0.972 & 0.977 & 0.955 \\
\textbf{Accessory pathway conduction} & 0.972 & 0.888 & 0.869 & 0.800 & 0.833 & 0.970 & 0.982 & 0.958 \\
\textbf{AV block} & 0.971 & 0.833 & 0.856 & 0.681 & 0.759 & 0.965 & 0.987 & 0.956 \\
\textbf{1st degree AV block} & 0.978 & 0.849 & 0.859 & 0.761 & 0.807 & 0.978 & 0.988 & 0.969 \\
\textbf{Bifascicular block} & 0.987 & 0.800 & 0.783 & 0.739 & 0.760 & 0.992 & 0.993 & 0.985 \\
\textbf{RBBB} & 0.987 & 0.909 & 0.879 & 0.821 & 0.849 & 0.983 & 0.989 & 0.975 \\
\textbf{LBBB} & 0.986 & 0.836 & 0.848 & 0.687 & 0.759 & 0.985 & 0.994 & 0.979 \\
\textbf{Myocardial infarction} & 0.928 & 0.820 & 0.812 & 0.661 & 0.729 & 0.879 & 0.942 & 0.864 \\
\textbf{Electronic pacemaker} & 0.976 & 0.812 & 0.863 & 0.763 & 0.810 & 0.988 & 0.994 & 0.983 \\
\hline
\end{tabular}
\end{table}

\begin{table}
\caption{\textbf{UHN-ECG reduced LVEF test results.} Label-specific test performances across all experiment suites using the full-scale training dataset.}
\label{tab:uhn_lvef}
\centering
\begin{tabular}{llcccccccc}
\hline
\textbf{Label} & \textbf{Experiment} & \textbf{AUROC} & \textbf{AUPRC} & \textbf{Recall} & \textbf{Precision} & \textbf{F1} & \textbf{Specificity} & \textbf{NPV} & \textbf{Accuracy} \\
\hline
\textbf{LVEF$\boldsymbol\leq$30\%} & \textbf{Full} & 0.918 & 0.674 & 0.678 & 0.594 & 0.633 & 0.914 & 0.938 & 0.876 \\
                                   & \textbf{Linear} & 0.905 & 0.652 & 0.703 & 0.565 & 0.626 & 0.899 & 0.942 & 0.868 \\
                                   & \textbf{Random Init.} & 0.875 & 0.585 & 0.647 & 0.490 & 0.558 & 0.874 & 0.930 & 0.839 \\
                                   & \textbf{Nejedly} & 0.891 & 0.623 & 0.691 & 0.522 & 0.595 & 0.882 & 0.939 & 0.852 \\
                                   & \textbf{SE-WRN} & 0.895 & 0.624 & 0.688 & 0.518 & 0.591 & 0.880 & 0.938 & 0.850 \\
\hline
\textbf{LVEF$\boldsymbol\leq$35\%} & \textbf{Full} & 0.921 & 0.754 & 0.758 & 0.659 & 0.705 & 0.892 & 0.931 & 0.864 \\
                                   & \textbf{Linear} & 0.905 & 0.723 & 0.775 & 0.602 & 0.678 & 0.860 & 0.933 & 0.842 \\
                                   & \textbf{Random Init.} & 0.876 & 0.660 & 0.723 & 0.544 & 0.621 & 0.834 & 0.917 & 0.810 \\
                                   & \textbf{Nejedly} & 0.893 & 0.696 & 0.755 & 0.586 & 0.660 & 0.854 & 0.927 & 0.832 \\
                                   & \textbf{SE-WRN} & 0.896 & 0.705 & 0.761 & 0.588 & 0.664 & 0.854 & 0.929 & 0.834 \\
\hline
\textbf{LVEF$\boldsymbol\leq$40\%} & \textbf{Full} & 0.929 & 0.825 & 0.880 & 0.643 & 0.743 & 0.824 & 0.950 & 0.839 \\
                                   & \textbf{Linear} & 0.914 & 0.785 & 0.889 & 0.596 & 0.714 & 0.784 & 0.951 & 0.811 \\
                                   & \textbf{Random Init.} & 0.879 & 0.713 & 0.870 & 0.533 & 0.661 & 0.726 & 0.939 & 0.764 \\
                                   & \textbf{Nejedly} & 0.896 & 0.754 & 0.865 & 0.576 & 0.692 & 0.771 & 0.941 & 0.796 \\
                                   & \textbf{SE-WRN} & 0.900 & 0.760 & 0.867 & 0.583 & 0.697 & 0.777 & 0.942 & 0.801 \\
\hline
\textbf{LVEF$\boldsymbol\leq$50\%} & \textbf{Full} & 0.911 & 0.878 & 0.913 & 0.666 & 0.770 & 0.701 & 0.925 & 0.785 \\
                                   & \textbf{Linear} & 0.889 & 0.849 & 0.894 & 0.646 & 0.750 & 0.681 & 0.908 & 0.765 \\
                                   & \textbf{Random Init.} & 0.860 & 0.800 & 0.910 & 0.590 & 0.716 & 0.588 & 0.909 & 0.715 \\
                                   & \textbf{Nejedly} & 0.875 & 0.827 & 0.910 & 0.600 & 0.723 & 0.604 & 0.911 & 0.725 \\
                                   & \textbf{SE-WRN} & 0.886 & 0.843 & 0.913 & 0.614 & 0.734 & 0.626 & 0.917 & 0.739 \\
\hline
\end{tabular}
\end{table}

\newpage

\subsection{Segment aggregation} \label{sec:aggregation}

Our interpretation tasks rely on annotated 10 s recordings, however our model evaluates each recording by cropping it into two non-overlapping 5 s segments, which become inputs utilizing the same label. This cropping is necessary to generate positive pairs for the CMSC contrastive objective. However, it means that we are predicting labels given partial information, where certain diagnoses may not even present in a given segment. To investigate how this affects our results, we explored aggregating model logits by taking the maximum, mean, and minimum for each two adjacent segments. \\

Label-specific aggregation methods are selected according to which yields the highest AUPRC on the validation set, where the best-performing methods tend to follow a common schema. Labels with 'max' are typically diagnosed using distinct features such as pacemaker spikes, ectopic beats, and artifacts. Aggregation may help here considerably when such markers are less discernible, or perhaps absent, in one of the segments. The 'mean' labels generally represent more continuous patterns such as rhythms and abnormalities resulting in a sustained change in presentation. Gains with this method are generally more modest and may be acting as an ensemble prediction which benefits from more complete information. Only \textit{Normal sinus rhythm} used the 'min' method, which is reasonable since this condition would not be stated unless both segments met the necessary criteria. \\

This aggregation experiment crudely quantifies how the model might have performed having seen the full ECG sample. The aggregated metrics may serve as a more accurate depiction of ECG-FM's capabilities, clarifying that certain poor performance is not due to inherent shortcomings as an encoder, but rather to an evaluative limitation. Although we find this aggregation methodology to be reasonable, we refrain from adopting it in the main paper to avoid over-complicating our method.

\begin{table}
\caption{\textbf{Segment-aggregated UHN-ECG interpretation results.} Segment-aggregated test results for \textit{Full}. Percentages indicate metric increases over the non-aggregated results.} \label{tab:aggregation}
\centering
\begin{tabular}{l|cclcl}
\hline
\textbf{Label} & \textbf{Method} & \textbf{AUROC} &  & \textbf{AUPRC} &  \\
\hline
\textbf{Poor data quality} & max & 0.924 & (+2.69\%) & 0.694 & (+5.54\%) \\
\textbf{Sinus rhythm} & mean & 0.993 & (+0.07\%) & 0.998 & (+0.03\%) \\
\textbf{Normal sinus rhythm} & min & 0.987 & (+0.87\%) & 0.985 & (+1.52\%) \\
\textbf{PVC} & max & 0.987 & (+4.61\%) & 0.877 & (+16.65\%) \\
\textbf{Tachycardia} & mean & 0.997 & (+0.08\%) & 0.986 & (+0.26\%) \\
\textbf{Ventricular tachycardia} & max & 0.997 & (+0.43\%) & 0.465 & (+2.68\%) \\
\textbf{SVT with aberrancy} & max & 0.987 & (+0.4\%) & 0.835 & (+3.39\%) \\
\textbf{Atrial fibrillation} & mean & 0.997 & (+0.06\%) & 0.973 & (+0.51\%) \\
\textbf{Atrial flutter} & mean & 0.987 & (+0.07\%) & 0.774 & (+1.77\%) \\
\textbf{Bradycardia} & mean & 0.997 & (+0.2\%) & 0.982 & (+1.22\%) \\
\textbf{Accessory pathway conduction} & mean & 0.996 & (+0.06\%) & 0.976 & (+0.34\%) \\
\textbf{AV block} & mean & 0.988 & (+0.12\%) & 0.915 & (+0.71\%) \\
\textbf{1st degree AV block} & mean & 0.993 & (+0.1\%) & 0.928 & (+0.84\%) \\
\textbf{2nd degree AV block} & mean & 0.987 & (+0.14\%) & 0.734 & (+2.18\%) \\
\textbf{Bifascicular block} & mean & 0.991 & (+0.05\%) & 0.704 & (+2.26\%) \\
\textbf{RBBB} & mean & 0.992 & (+0.07\%) & 0.942 & (+0.33\%) \\
\textbf{LBBB} & mean & 0.993 & (+0.06\%) & 0.869 & (+0.5\%) \\
\textbf{Myocardial infarction} & mean & 0.941 & (+0.25\%) & 0.825 & (+0.64\%) \\
\textbf{Electronic pacemaker} & max & 0.991 & (+0.81\%) & 0.948 & (+1.86\%) \\
\textbf{Ventricular pacing} & max & 0.998 & (+0.16\%) & 0.977 & (+0.78\%) \\
\textbf{Atrial pacing} & max & 0.996 & (+0.06\%) & 0.929 & (+0.91\%) \\
\hline
\end{tabular}
\end{table}

\newpage

\subsection{Patient demographics} \label{sec:demographics}

\begin{table}
\centering
\caption{\textbf{Biological age.} Shown is the Mean $\pm$ STD of biological age in years across UHN-ECG's downsteam task manifests.}
\label{tab:age}
\begin{tabular}{l|cc}
\hline
\textbf{Split} & \textbf{Interpretation/Abnormal cTn} & \textbf{Reduced LVEF} \\
\hline
\textbf{Train} & $62.2 \pm 19.5$ & $65.9 \pm 16.8$ \\
\textbf{Valid} & $60.8 \pm 19.3$ & $66.3 \pm 16.1$ \\
\textbf{Test}  & $59.8 \pm 19.4$ & $65.6 \pm 16.5$ \\
\hline
\end{tabular}
\end{table}

\begin{table}
\centering
\caption{\textbf{Biological sex.} Shown is distribution of sex (\% Female) across UHN-ECG's downsteam task manifests.}
\label{tab:bio_sex}
\begin{tabular}{l|cc}
\hline
\textbf{Split} & \textbf{Interpretation/Abnormal cTn} & \textbf{Reduced LVEF} \\
\hline
\textbf{Train} & 45.7\% & 37.3\% \\
\textbf{Valid} & 45.9\% & 39.6\% \\
\textbf{Test}  & 44.9\% & 40.2\% \\
\hline
\end{tabular}
\end{table}

\subsection{Interpretation Text Parsing} \label{sec:interpretation-parsing}
In clinical care settings, ECG interpretations are often recorded as free-text. While this approach is convenient and allows for unbounded precision, it can be challenging to translate this unstructured format into precise labels which can be digested by AI models. Synonyms, acronyms, grammar, typographical errors, evolving medical terminology, and implied findings are all examples of complexities which, if not handled with care, can severely lessen label quality, which in turn reduces model effectiveness and evaluative correctness. \\

Maintaining positional information, we applied pattern matching which was manually curated to parse free-text and match over $99\%$ of UHN-ECG interpretations completely. Derived from these patterns are a series of entities (e.g., 'tachycardia', 'infarction'), descriptors (e.g., 'probably', 'moderate', 'acute'), and connectives (e.g., 'associated with', 'transitions to'). Relevant information from the descriptors and connectives are distilled down into their corresponding entities. We map the resulting entities into labels which can be flexibly manipulated. Using clinician-in-the-loop decision making, we constructed a knowledge graph encoding label relationships which are true by definition. We used it to recursively mark labels as true, for example, labeling \textit{Ventricular tachycardia} when \textit{Torsades de Pointes}, one form of polymorphic ventricular tachycardia, was specifically stated. Without this component, we suspect that the model would learn physiologically arbitrary distinctions which would prove counterproductive to accurate interpretation.

\subsection{Performance curves} \label{sec:curves}
For all tasks, we show receiver operating characteristic (ROC) curves and the precision-recall curves (PRC) for each label in our \textit{Full}, \textit{Random Init.}, and \textit{Linear} experiment suites, reporting the label names in the legend.

\begin{figure}
    \centering
    \includegraphics[scale=0.45]{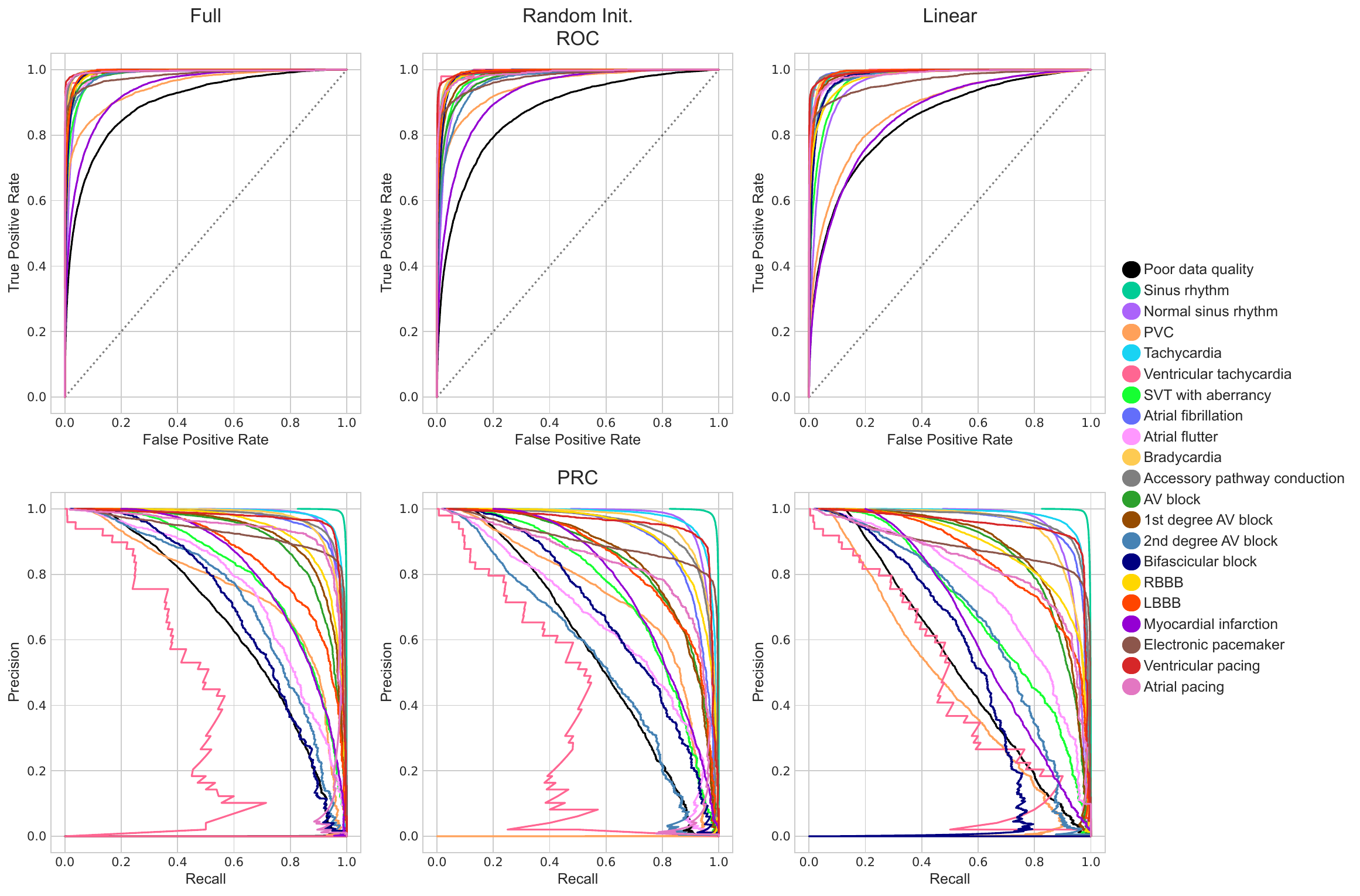}
    \caption{\textbf{UHN-ECG interpretation task test performance curves.}}
    \label{fig:curves_uhn_interpretation}
\end{figure}

\begin{figure}
    \centering
    \includegraphics[scale=0.45]{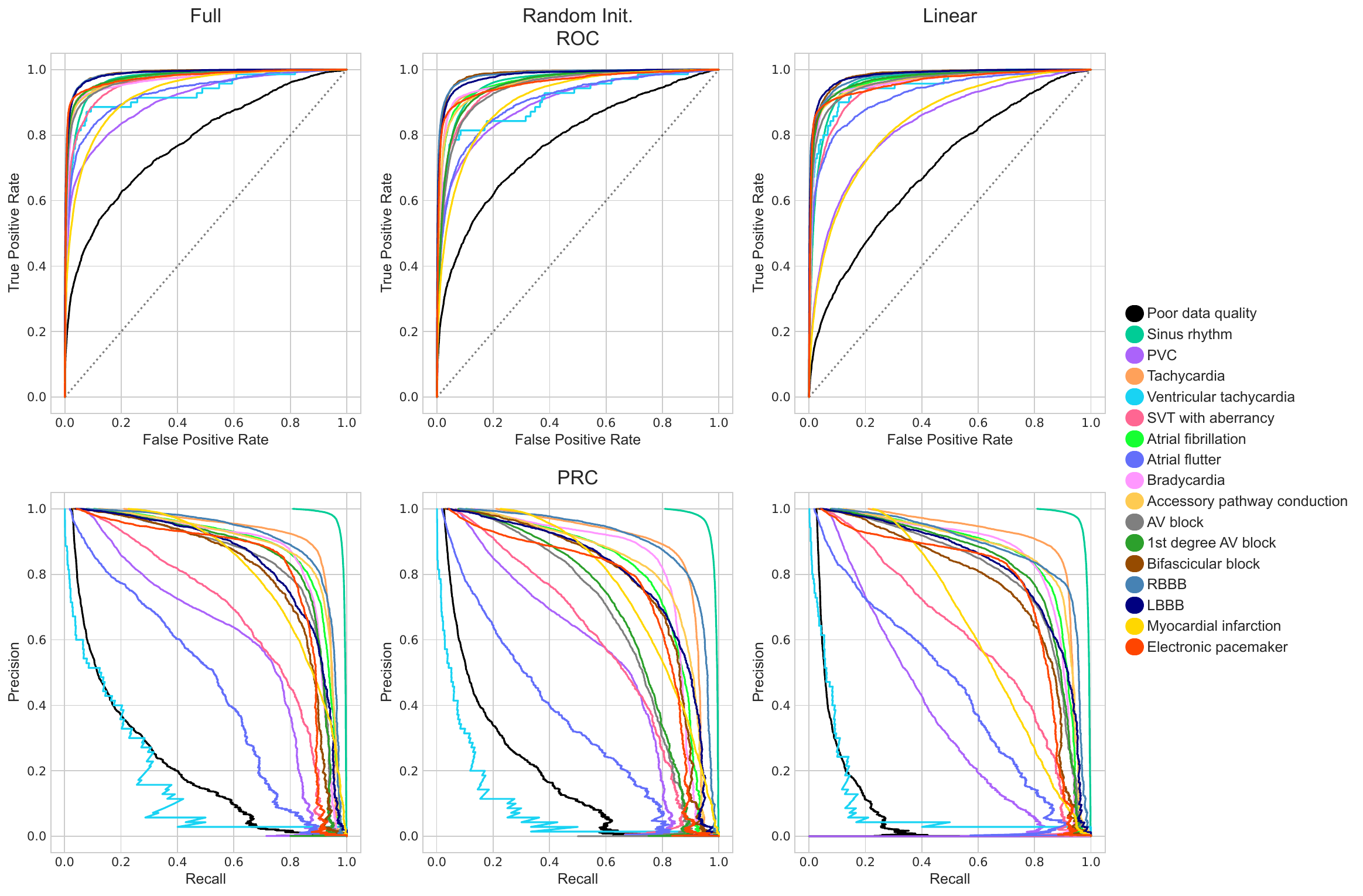}
    \caption{\textbf{MIMIC-IV-ECG machine reads task test performance curves.}}
    \label{curves_mimic_iv_ecg_machine_reads:}
\end{figure}

\begin{figure}
    \centering
    \includegraphics[scale=0.45]{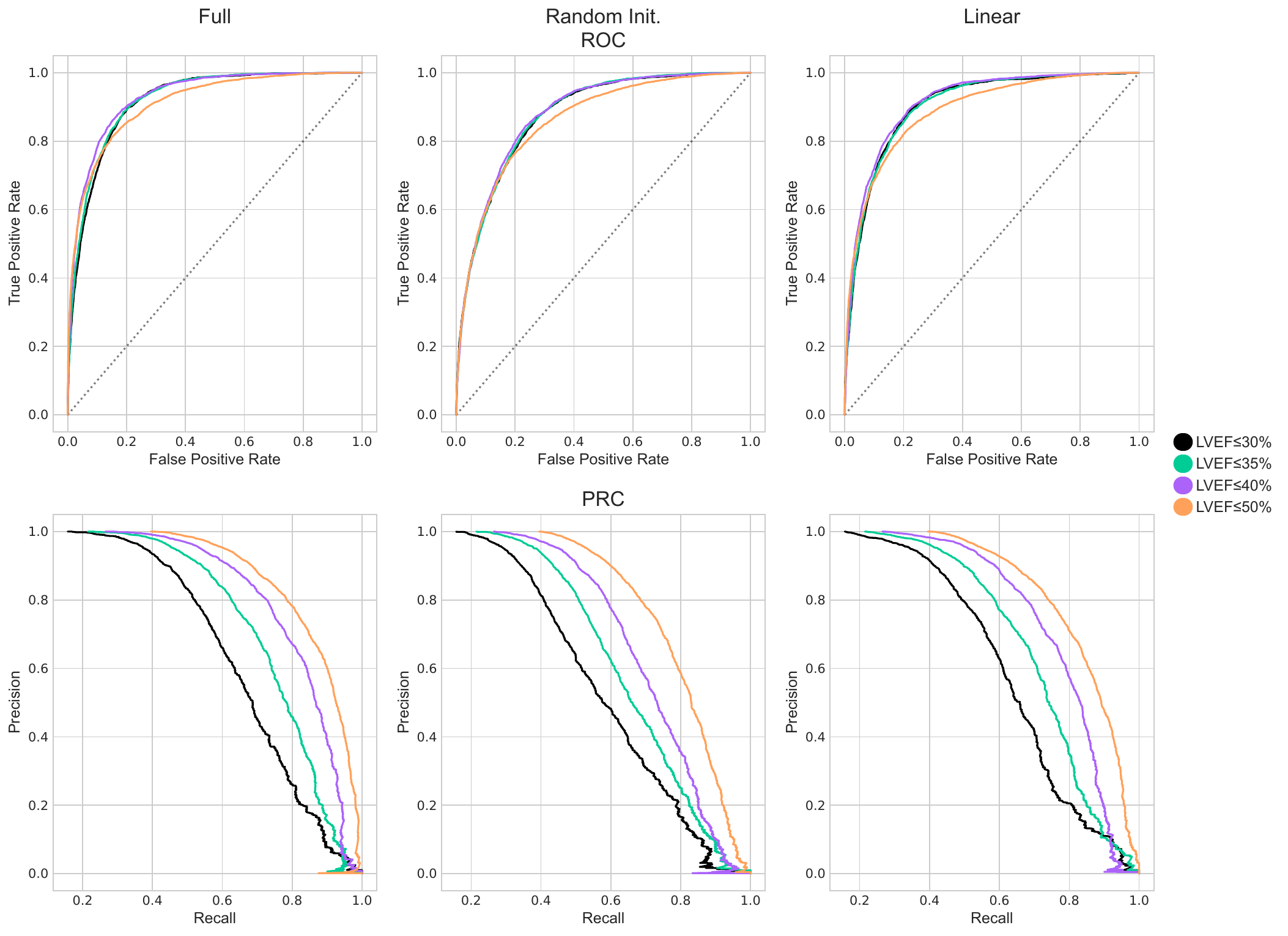}
    \caption{\textbf{UHN-ECG reduced LVEF task test performance curves.}}
    \label{fig:curves_uhn_lvef}
\end{figure}

\end{document}